\documentclass[10pt,twocolumn,letterpaper]{article}

\usepackage{iccv}
\usepackage{times}
\usepackage{epsfig}
\usepackage{graphicx}
\usepackage{amsmath}
\usepackage{amssymb}

\usepackage{mathtools}
\DeclarePairedDelimiter{\floor}{\lfloor}{\rfloor}
\DeclarePairedDelimiter{\ceil}{\lceil}{\rceil}
\usepackage{multirow}

\usepackage[dvipsnames]{xcolor}
\usepackage{kotex}
\usepackage{booktabs}

\usepackage{subfig}
\usepackage{tabularx}

\usepackage{varwidth}

\usepackage{textcomp} 

\usepackage{floatrow}
\newfloatcommand{capbtabbox}{table}[][\FBwidth]
\usepackage{xcolor}
\usepackage{pifont}

\usepackage{dsfont}
\usepackage{bbm}
\newcommand{\1}[1]{\mathbbm{1}\left[#1\right]}

\usepackage{algorithm,algorithmic}

\renewcommand{\algorithmiccomment}[1]{\bgroup\hfill\small\textsl{#1}\egroup}

\DeclareMathOperator*{\argmax}{arg\,max}
\DeclareMathOperator*{\argmin}{arg\,min}

\newcommand{\Fref}[1]{Fig.~\ref{#1}}
\newcommand{\Sref}[1]{Sec.~\ref{#1}}
\newcommand{\Tref}[1]{Table~\ref{#1}}
\newcommand{\Aref}[1]{Algorithm~\ref{#1}}

\newcommand{\R}{\mathbb{R}}

\def\@fnsymbol#1{\ensuremath{\ifcase#1\or *\or \dagger\or \ddagger\or
   \mathsection\or \mathparagraph\or \|\or **\or \dagger\dagger
   \or \ddagger\ddagger \else\@ctrerr\fi}}

\newcommand{\ssymbol}[1]{^{\@fnsymbol{#1}}}

\newcommand*\colourcheck[1]{%
  \expandafter\newcommand\csname #1check\endcsname{\textcolor{#1}{\ding{51}}}%
}
\colourcheck{ForestGreen}

\newcommand*\colourx[1]{%
  \expandafter\newcommand\csname #1x\endcsname{\textcolor{#1}{\ding{55}}}%
}
\colourx{blue}
\colourx{green}
\colourx{red}

\usepackage[pagebackref=true,breaklinks=true,letterpaper=true,colorlinks,bookmarks=false]{hyperref}

\iccvfinalcopy 

\def\iccvPaperID{1186} 

\ificcvfinal\fi

\begin{document}

\title{\vspace{-1.em}Learning Action Completeness from Points\\for Weakly-supervised Temporal Action Localization\vspace{-.5em}}

\author{Pilhyeon Lee\textsuperscript{\rm 1} \qquad
    Hyeran Byun\textsuperscript{\rm 1,2}\thanks{Corresponding author} \vspace{.5em} \\
\textsuperscript{\rm 1}Department of Computer Science, Yonsei University\\
\textsuperscript{\rm 2}Graduate school of AI, Yonsei University\\
{\tt\small \{lph1114, hrbyun\}@yonsei.ac.kr}
}

\maketitle
\ificcvfinal\thispagestyle{empty}\fi

\begin{abstract}
We tackle the problem of localizing temporal intervals of actions with only a single frame label for each action instance for training. Owing to label sparsity, existing work fails to learn action completeness, resulting in fragmentary action predictions. In this paper, we propose a novel framework, where dense pseudo-labels are generated to provide completeness guidance for the model. Concretely, we first select pseudo background points to supplement point-level action labels. Then, by taking the points as seeds, we search for the optimal sequence that is likely to contain complete action instances while agreeing with the seeds. To learn completeness from the obtained sequence, we introduce two novel losses that contrast action instances with background ones in terms of action score and feature similarity, respectively. Experimental results demonstrate that our completeness guidance indeed helps the model to locate complete action instances, leading to large performance gains especially under high IoU thresholds. Moreover, we demonstrate the superiority of our method over existing state-of-the-art methods on four benchmarks: THUMOS'14, GTEA, BEOID, and ActivityNet.
Notably, our method even performs comparably to recent fully-supervised methods, at the 6$\times$ cheaper annotation cost.
Our code is available at  \url{https://github.com/Pilhyeon}.
\end{abstract}

\section{Introduction}
\label{sec:intro}

\begin{figure}[t]
  \centering
  \includegraphics[clip=true, width=0.95\textwidth]{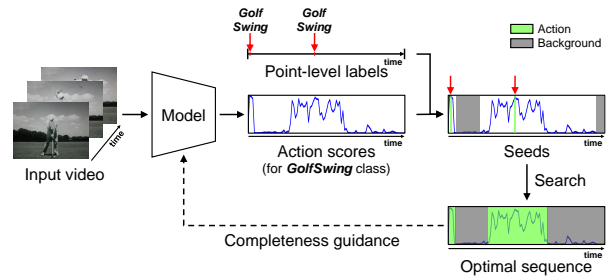}
  \caption{Simplified illustration of our idea.
    We use points as seeds to find the optimal sequence, which in turn provides completeness guidance to the model.
    }
  \label{fig:intro_fig}
\end{figure}

The goal of temporal action localization lies in locating starting and ending timestamps of action instances and classifying them.
Thanks to the various applications~\cite{Ma2005AGF,Vishwakarma2012ASO,Xiong2019LessIM}, it has drawn much attention from researchers, leading to the rapid and remarkable progress in the fully-supervised setting (\ie, frame-level labels)~\cite{Liu2019MGG,shou2017cdc,shou2016temporal,xu2017r}.
Meanwhile, there appear attempts to reduce the prohibitively expensive cost of annotating individual frames by devising weakly-supervised models with video-level labels~\cite{Fernando2020WSGN,Ma_2021_ASL,wang2017untrimmednets,Yuan2019MARGINALIZEDAA}.
However, they fall largely behind the fully-supervised counterparts, mainly on account of their weak ability to distinguish action and background frames~\cite{lee2020background,lee2021um,Nguyen2019WeaklySupervisedAL,Xu2019SegregatedTA}.

To narrow the performance gap between them, another level of weak supervision has been proposed recently, namely the point-supervised setting.
In this setting, only a single timestamp (point) with its action category is annotated for each action instance during training.
In terms of the labeling cost, point-level labels require a negligible extra cost compared to video-level ones, while being $6\times$ cheaper than frame-level ones (50\textit{s} \textit{vs.} 300\textit{s} per 1-min video)~\cite{ma2020sfnet}.

Despite the affordable cost, it offers coarse locations as well as the total number of action instances, thus bringing a strong ability in spotting actions to the models.
Consequently, point-supervised methods show comparable or even superior performances to fully-supervised counterparts under low intersection over union (IoU) thresholds.
However, it has been revealed that they suffer from incomplete predictions, resulting in highly inferior performances in the case of high IoU thresholds.
We conjecture that this problem is attributed to the sparse nature of point-level labels that induces the models to learn only a small part of actions rather than the full extent of action instances.
In other words, they fail to learn \textit{action completeness} from the point annotations.
Although SF-Net~\cite{ma2020sfnet} mines pseudo action and background points to alleviate the label sparsity, they are discontinuous and thus do not provide completeness cues.

In this paper, we aim to allow the model to learn action completeness under the point-supervised setting.
To this end, we introduce a new framework, where dense pseudo-labels (\ie, sequences) are generated based on the point annotations to provide completeness guidance to the model.
The overall workflow is illustrated in \Fref{fig:intro_fig}.

Technically, we first select pseudo background points to augment point-level action labels.
As aforementioned, such point annotations are discontiguous, so it is infeasible to learn completeness from them.
To that end, we propose to search for the optimal sequence covering complete action instances among candidates consistent with the point labels.
However, it is non-trivial to measure how complete the instances in each candidate sequence are, without full supervision.
To realize it, we borrow the outer-inner-contrast concept~\cite{shou2018autoloc} as a proxy for instance completeness.
Intuitively, a complete action instance generally shows large score contrast, \ie, much higher action scores for inner frames than those for surrounding frames.
In contrast, a fragmentary instance probably has high action scores in its outer region (still within the action), leading to small score contrast.
This can be generalized for background instances as well.
Based on this property, we derive the score of an input sequence by aggregating the score contrast of action and background instances constituting the sequence.
By maximizing the score, we can obtain the optimal sequence that is likely to be well-aligned with the ground-truth we do not have.
In experiments, we present the accuracy of optimal sequences and the correlation between score contrast and completeness.

From the obtained sequence, the model is supposed to learn action completeness.
To this end, we design score contrastive loss to maximize the agreement between the model outputs and the optimal sequence, by enlarging the completeness of the sequence.
With the loss, the model is trained to discriminate each action (background) instance from its surroundings in terms of action scores.
Moreover, we introduce feature contrastive loss to encourage feature discrepancy between action and background instances.
Experiments validate that the proposed losses complementarily help the model to detect complete action instances, leading to large performance gains under high IoU thresholds.

To summarize, our contributions are three-fold.
    
\begin{itemize}
    \item We introduce a new framework, where the dense optimal sequence is generated to provide completeness guidance to the model in the point-supervised setting.
    \item We propose two novel losses that facilitate the action completeness learning by contrasting action instances with background ones with respect to action score and feature similarity, respectively.
    \item Our model achieves a new state-of-the-art with a large gap on four benchmarks. Furthermore, it even performs favorably against fully-supervised approaches.
\end{itemize}

\begin{figure*}[t]
  \centering
  \includegraphics[clip=true, width=0.97\textwidth]{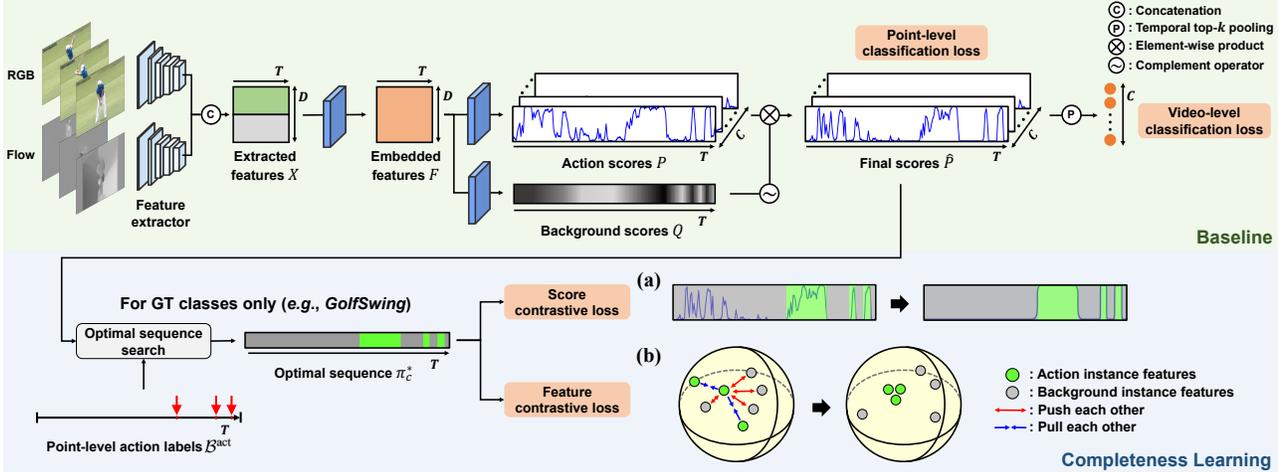}
  \caption{Overview of the proposed method. Besides the conventional objectives, \ie, video-level and point-level classification losses, we propose to learn action completeness (the lower part).
  Based on the final action scores, the optimal sequence is selected among candidates consistent with the point-level labels.
  It in turn provides completeness guidance with two proposed losses that contrast action instances with background ones with respect to (a) action score and (b) feature similarity.}
  \label{fig:main_architecture}
\end{figure*}

\section{Related Work}
\label{sec:related_work}
\noindent\textbf{Fully-supervised temporal action localization.}~~
In order to tackle temporal action localization, fully-supervised methods rely on precise temporal annotations, \ie, \textbf{frame-level labels}.
They mainly adopt the two-stage paradigm (proposal generation and classification), and can be roughly categorized into two groups regarding the way to generate proposals.
The first group prepares a large number of proposals using the sliding window technique~\cite{chao2018rethinking,shou2017cdc,shou2016temporal,xiong2017pursuit,yang2018exploring,yuan2016temporal,zhao2017temporal}.
On the other hand, the second group first predicts the probability of each frame being a start (end) point of an action instance, and then uses the combinations of probable start and end points as proposals~\cite{lin2020fast,Lin2019BMNBN,lin2018bsn,zhao2020bottom-up}.
Meanwhile, there are graph modeling methods taking snippets~\cite{bai2020bcgnn,xu2020g-tad} or proposals~\cite{zeng2019p-gcn} as nodes.
Different from fully-supervised methods that utilize expensive frame-level labels for action completeness learning, our method enables it with only point-level labels by introducing a novel framework.

\noindent\textbf{Weakly-supervised temporal action localization.}~~
To alleviate the cost issue of frame-level labels, many attempts have been made recently to solve the same task in the weakly-supervised setting, mainly using \textbf{video-level labels}.
Untrimmednets~\cite{wang2017untrimmednets} tackle it by selecting segments that contribute to video-level classification.
STPN~\cite{nguyen2018weakly} puts a constraint that key frames should be sparse.
In addition, there are background modeling approaches under the video-supervised setting~\cite{Islam2021HAM-Net,lee2020background,lee2021um,Nguyen2019WeaklySupervisedAL}.
To learn reliable attention weights, DGAM~\cite{Shi2020DGAM} designs a generative modeling, while EM-MIL~\cite{luo2020EMMIL} adopts the Expectation-maximization strategy.
Meanwhile, metric learning is utilized for action representation learning~\cite{Islam2020metric,Narayan20193CNetCC,paul2018w} or action-background separation~\cite{min2020A2CL}.
There are also methods that explore sub-actions~\cite{Jain2020ActionBytes,Luo_2021_AUMN} or exploit the complementarity of RGB and flow modalities~\cite{Yang_2021_UGCT,zhai2020TSCN}.
Besides, several methods leverage external information, \eg, action count~\cite{Narayan20193CNetCC,Xu2019SegregatedTA}, pose~\cite{zhang2020MultiinstanceMA} or audio~\cite{Lee2021audio-visual}.
Moreover, some approaches aim to detect complete action instances by aggregating multiple predictions~\cite{liu2019completeness}, erasing the most discriminative part~\cite{singh2017hide,zhong2018step}, or directly regressing the action intervals~\cite{Liu2019WeaklyST,shou2018autoloc}.

Most recently, \textbf{point-level supervision} starts to be explored, which provides rich information at an affordable cost.
Moltisanti~\etal~\cite{Moltisanti2019CVPR} first utilize the point-level labels for action localization.
SF-Net~\cite{ma2020sfnet} adopts the pseudo label mining strategy to acquire more labeled frames.
Meanwhile, Ju~\etal~\cite{ju2020point} perform boundary regression based on key frame prediction.
However, they do not explicitly consider action completeness, and therefore produce predictions that cover only part of action instances.
In contrast, we propose to learn action completeness from dense pseudo-labels by contrasting action instances with surrounding background ones.
In \Sref{sec:experiments}, the efficacy of our method is clearly verified
with notable performance boosts at high IoU thresholds.

\section{Method}
\label{sec:method}
In this section, we first describe the problem setting and detail the baseline setup.
Afterward, the optimal sequence search is elaborated, followed by our action completeness learning strategy.
Lastly, we explain the joint learning and the inference of our model.
The overall architecture of our method is illustrated in \Fref{fig:main_architecture}.

\noindent\textbf{Problem setting.}
Following~\cite{ju2020point,ma2020sfnet}, we set up the problem of point-supervised temporal action localization.
Given an input video, a single point and the category for each action instance is provided, \ie, $\mathcal{B}^\text{act}=\{(t_{i}, {\mathbf{y}}_{t_{i}})\}_{i=1}^{M^\text{act}}$, where the $i$-th action instance is labeled at the $t_i$-th segment (frame) with its action label $\mathbf{y}_{t_{i}}$, and $M^\text{act}$ is the total number of action instances in the input video.
The points are sorted in temporal order (\ie, $t_i<t_{i+1}$).
The label ${\mathbf{y}}_{t_{i}}$ is a binary vector with ${y}_{t_{i}}[c]=1$ if the $i$-th action instance contains the $c$-th action class and otherwise $0$ for $C$ action classes.
It is worth noting that the video-level label $\mathbf{y}^{\text{vid}}$ can be readily acquired by aggregating the point-level ones, \ie, $y^{\text{vid}}[c]=\1{\sum_{i=1}^{M^{\text{act}}}{y_{t_{i}}[c]}>0}$, where $\1{\cdot}$ is the indicator function.

\subsection{Baseline Setup}
\label{subsec:baseline}

Our baseline is shown in the upper part of \Fref{fig:main_architecture}.
We first divide the input video into 16-frame segments, which are then fed to the pre-trained feature extractor.
Following~\cite{lee2020background,paul2018w}, we exploit both of RGB and flow streams with early-fusion.
The two-stream features are fused by concatenation, resulting in $X\in\R^{D\times T}$, where $D$ and $T$ denote the feature dimension and the number of segments, respectively.

The extracted features then go through a single 1D convolutional layer followed by ReLU activation, which produces the embedded features $F$.
In practice, we set the dimension of the embedded features to the same as that of the extracted features $X$, \ie, $F\in\R^{D\times T}$.
Afterward, the embedded features are fed into a 1D convolutional layer with the sigmoid function, to predict the segment-level class scores $P\in\R^{C\times T}$, where $C$ indicates the number of action classes.
Meanwhile, we derive the class-agnostic background scores $Q\in\R^{T}$, to model background frames which do not belong to any action classes.
Thereafter, we fuse the action scores with the complement of background probability to get the final scores $\hat{P}$, \ie, $\hat{p}_{t}[c]=p_{t}[c](1-q_{t})$.
This fusion strategy is similar to that of \cite{lee2021um}, although the out-of-distribution modeling is not incorporated in our model.

The segment-level action scores are then aggregated to build a single video-level class score.
We use the temporal top-$k$ pooling for aggregation as in~\cite{lee2020background,paul2018w}.
Formally, the video-level probability is calculated as follows.
\begin{equation}
    \hat{p}^{\text{vid}}[c]=\frac{1}{k}\max_{\substack{S\subset \hat{P}[c,:]}}{\sum_{\forall m \in S}{m}},
\end{equation}
where $k=\floor{\frac{T}{8}}$ and $S$ denotes all possible subsets of $\hat{P}[c,:]$ containing $k$ segments, \ie, $|S|=k$.

Our baseline model includes two loss functions using video- and point-level labels respectively.
As aforementioned, the video-level class label $y^{\text{vid}}[c]$ can be derived by accumulating the point-level labels. 
The video-level classification loss is then calculated with binary cross-entropy.
\begin{equation}
\begin{aligned}
    \mathcal{L}_{\text{video}}=-\sum_{c=1}^{C}\Big(&y^{\text{vid}}[c]\log{\hat{p}^{\text{vid}}[c]]}\\ &+(1-y^{\text{vid}}[c])\log{(1-\hat{p}^{\text{vid}}[c])}\Big).
\end{aligned}
\label{eq:video_loss}
\end{equation}

The point-level classification loss is also computed by binary cross-entropy but involving the background term for effectively training $Q$.
In addition, we adopt the focal loss~\cite{lin2017focal} to facilitate the training process.
Formally, the classification loss for action points is defined as follows.
\begin{equation}
\resizebox{1.0\hsize}{!}{
$
\begin{aligned}
    \mathcal{L}_{\text{point}}^{\text{act}}=&-\frac{1}{M^{\text{act}}}\sum_{\forall(t, \mathbf{y}_{t})\in\mathcal{B}^{\text{act}}}\bigg(\sum_{c=1}^{C}\Big(y_{t}[c]{(1-\hat{p}_{t}[c])}^\beta\log{\hat{p}_{t}[c]}\\
    &+(1-y_{t}[c]){\hat{p}_{t}[c]}^\beta\log{(1-\hat{p}_{t}[c])}\Big)+q_{t}^{\beta}\log{(1-q_t)}\bigg),
\end{aligned}
$
}
\label{eq:point_act}
\end{equation}
where $M^{\text{act}}$ indicates the number of action instances in the video and $\beta$ is the focusing parameter, which is set to $2$ following the original paper~\cite{lin2017focal}.

Training only with action points would lead the network to always produce low background scores rather than learn to separate action and background.
Therefore, we gather some pseudo background points to supplement action ones.
Our principle for selection is that at least one background frame must be placed between two adjacent action instances to separate them.
By the problem definition, two different action points are sampled from different instances, so we use the action points as surrogates for the corresponding instances.
Concretely, between two adjacent action points, we find the segments whose background scores $q_t$ are larger than the threshold $\gamma$.
If no segment satisfies the condition in a section, we select one with the largest background score.
Meanwhile, for the case where multiple background points are selected in a section, we mark all points between them as background, since it is trivial that no action exists there.
In practice, this strategy is shown to be more effective than global mining~\cite{ma2020sfnet} by collecting more hard points.
Given the pseudo background point set, $\mathcal{B}^\text{bkg}=\{t_{j}\}_{j=1}^{M^\text{bkg}}$, the classification loss for background points is computed by:

\begin{equation}
\resizebox{1.0\hsize}{!}{
$
\begin{aligned}
    \mathcal{L}_{\text{point}}^{\text{bkg}}=-\frac{1}{M^{\text{bkg}}}\sum_{\forall t\in\mathcal{B}^{\text{bkg}}}\bigg(\sum_{c=1}^{C} {\hat{p}_{t}[c]}^\beta\log{(1-\hat{p}_{t}[c])}
    +(1-q_t)^\beta\log{q_t}\bigg),
\end{aligned}
$
}
\end{equation}
where $M^{\text{bkg}}$ denotes the number of the selected background points and $\beta$ is the focusing factor, the same with \eqref{eq:point_act}.
For pseudo background points, we penalize the final scores for all action classes, while encouraging the background scores.

The total point-level loss function is defined as the sum of the losses for action and pseudo background points.
\begin{equation}
    \mathcal{L}_{\text{point}}=\mathcal{L}_{\text{point}}^{\text{act}}+\mathcal{L}_{\text{point}}^{\text{bkg}}.
\label{eq:point_loss}
\end{equation}

\begin{figure*}[t]
  \centering
  \includegraphics[clip=true, width=0.95\textwidth]{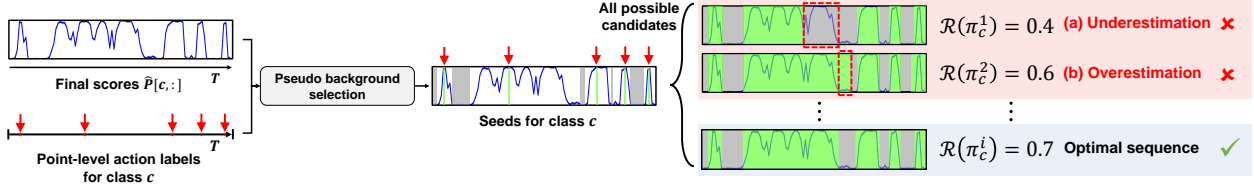}
  \caption{Optimal sequence search for class $c$. Given the final scores and the point-level labels, we select pseudo background points.
  Then, among all possible candidates, we search for the optimal sequence that maximizes the completeness score \eqref{eq:score}.}
  \label{fig:search_fig}
\end{figure*}

\subsection{Optimal Sequence Search}
\label{subsec:guidance}
As discussed in \Sref{sec:intro}, the point-level classification loss is insufficient to learn action completeness, as point labels cover only a small portion of action instances.
Therefore, we propose to generate dense pseudo-labels that can offer some hints about action completeness for the model.
In detail, we consider all possible sequence candidates consistent with the action and pseudo background points.
Among them, we find the optimal sequence that can provide good completeness guidance to the model.
However, it is non-trivial without full supervision to measure how well a candidate sequence covers complete action instances.
To enable it, we re-purpose the outer-inner-contrast concept~\cite{shou2018autoloc} as a proxy for judging the completeness score of a sequence.
Intuitively, the contrast between inner and outer scores is likely to be large for a complete action instance but small for a fragmentary one.
Note that our purpose is different from the original paper~\cite{shou2018autoloc}.
It was originally designed for parametric boundary regression.
In contrast, we utilize it as a scoring function to search for the optimal sequence, from which the model could learn action completeness.

Before detailing the scoring function, we present the formulation of candidate sequences.
Due to the multi-label nature of temporal action localization, we consider class-specific sequences for each action class.
Note that all segments belonging to other action classes are considered background for sequences of class $c$.
Then, a sequence is defined as multiple action and background (including other actions) instances that alternate consecutively.
Formally, a sequence of class $c$ can be expressed as $\pi_c=\{(s_{n}^{c}, e_{n}^{c}, z_{n}^{c})\}_{n=1}^{N_c}$, where $s_{n}^{c}$ and $e_{n}^{c}$ denote the start and end points of the $n$-th instance, respectively, while $N_c$ is the total number of instances for class $c$.
In addition, $z_{n}^{c}\in\{0,1\}$ indicates the type of the instance, \ie, $z_{n}^{c}=1$ if $n$-th instance is of the $c$-th action class, otherwise $0$ (background).

Given an input sequence, we compute its completeness score by averaging the contrast scores of individual action and background instances contained in the sequence.
It would be noted that the contrast scores of background instances are included in the calculation, which proves to be effective for finding more accurate optimal sequences, as will be shown in \Sref{subsec:analysis}.
Formally, the completeness score of a sequence $\pi_c$ for the $c$-th action class is computed by:
\begin{equation}
\resizebox{1.0\hsize}{!}{
$
\begin{aligned}
    \mathcal{R}(\pi_c)&=\frac{1}{N_c}\sum_{n=1}^{N_c}\bigg(\underbrace{\frac{1}{l_n^c}\sum_{t=s_{n}^{c}}^{e_{n}^{c}}{u_{n}^{c}(t)}}_{\text{Inner score}} \\
    &-\underbrace{\frac{1}{\ceil{\delta l_n^c}+\floor{\delta l_n^c}}\Big(\sum_{t=s_{n}^{c}-\ceil{\delta l_{n}^{c}}}^{s_{n}^{c}-1}u_{n}^{c}(t)+\sum_{t=e_{n}^{c}+1}^{e_{n}^{c}+\floor{\delta l_{n}^{c}}}u_{n}^{c}(t)}_{\text{Outer score}}\Big)\bigg), \\
    &\text{where}~~u_{n}^{c}(t)=
    \begin{cases}
        \hat{p}_t[c], & \text{if $z_{n}^{c}=1$}. \\
        1 - \hat{p}_t[c], & \text{otherwise}.
    \end{cases}
    ,
\end{aligned}
$
}
\label{eq:score}
\end{equation}
$l_{n}^{c}=e_{n}^{c}-s_{n}^{c}+1$ is the temporal length of the $n$-th instance of $\pi_c$, $\delta$ is a hyper-parameter adjusting the outer range (set to 0.25), and $N_c$ is the total number of action and background instances for class $c$.
Then, the optimal sequence for class $c$ can be obtained by finding the sequence that maximizes the score, \ie,
$\pi_c^*=\argmax_{\pi_c}\mathcal{R}(\pi_c)$ using \eqref{eq:score}.
The optimal sequence search process is illustrated in \Fref{fig:search_fig}.
By evaluating the completeness score, our method can reject underestimation (\Fref{fig:search_fig}a) and overestimation (\Fref{fig:search_fig}b) cases.
Consequently, we obtain the optimal sequence that is most likely to contain complete action instances.

However, the search space grows exponentially as $T$ increases, leading to the exorbitant cost for optimal sequence search.
To relieve the issue, we implement the search process with a greedy algorithm under a limited budget,
which results in greatly saving the computational cost.
Detailed algorithm and cost analysis are presented in Sec.~\textcolor{red}{B} of the appendix.
Note that optimal sequence search is performed only for the action classes contained in the video.

\subsection{Action Completeness Learning}
\label{subsec:completeness}
Given the class-specific optimal sequences $\{\pi_c^*\}_{c=1}^{C}$, our goal is to let the model learn action completeness.
To this end, we design two losses that enable completeness learning by contrasting action instances from background ones.
This helps in complete action predictions, as validated in \Sref{sec:experiments}.

Firstly, we propose \textit{score contrastive loss} that encourages the model to separate action (background) instances from their surroundings in terms of final scores.
It can be also interpreted as fitting the model outputs to the optimal sequences (\Fref{fig:main_architecture}a).
Formally, the loss is computed by:
\begin{equation}
    \mathcal{L}_{\text{score}}=\frac{1}{\sum_{c=1}^{C}y^{\text{vid}}[c]}\sum_{c=1}^{C}y^{\text{vid}}[c]\big(1-\mathcal{R}(\pi_c^*)\big)^\beta,
\label{eq:score_loss}
\end{equation}
where we use $\beta$-squared term to focus on the instances that are largely inconsistent with the optimal sequence ($\beta=2$).

Secondly, inspired by the recent success of contrastive learning~\cite{chen2020simCLR,he2020moco,khosla2020SupCon}, we design \textit{feature contrastive loss}.
Our intuition is that features from different instances but with the same action class should be closer to each other than any other background instances in the same video~(\Fref{fig:main_architecture}b).
We note that our loss differs from \cite{chen2020simCLR,he2020moco,khosla2020SupCon} in that they pull different views of an input image, whereas ours attracts different action instances in a given video.
In addition, ours does not need negative sampling from different images, as background instances are obtained from the same video.

To extract the representative feature for each action (or background) instance, we modify the segment of interest~(SOI) pooling~\cite{chao2018rethinking} by replacing max-pooling with random sampling.
In detail, we evenly divide each input instance into three intervals, from each of which a single segment is randomly sampled.
Then, the embedded features of the sampled segments are averaged, producing the representative feature $f_n^c$ for the $n$-th instance of the sequence $\pi_c^*$.

Taking the normalized instance features $\bar{f}_n^c$ as inputs, we derive feature contrastive loss.
The loss is computed only for the classes whose action counts are larger than 1, \ie, at least two action instances exist in the video.
Note that background instances do not attract each other.
Given the optimal sequences $\big\{\pi_{c}^{*}=\{(s_{n}^{c},e_{n}^{c},z_{n}^{c})\}_{n=1}^{N_c}\big\}_{c=1}^{C}$, the proposed feature contrastive loss is formulated as:
\begin{equation}
\resizebox{1.0\hsize}{!}{
$
\begin{aligned}
    \mathcal{L}_{\text{feat}}&=\frac{1}{\sum_{c=1}^{C}\1{\sum_{n=1}^{N_c}z_{n}^{c}>1}}\sum_{c=1}^{C}\1{\sum_{n=1}^{N_c}z_{n}^{c}>1}\ell_{\text{feat}}^c, \\
    \ell_{\text{feat}}^c&=-\frac{1}{\sum_{n=1}^{N_c}z_{n}^{c}}\sum_{n=1}^{N_c}z_{n}^{c}\log{\frac{\sum_{\forall o \neq n} z_{o}^{c}\text{exp}(\bar{f_n^c} \cdot \bar{f_o^c} / \tau)}{\sum_{\forall m \neq n} \text{exp}(\bar{f_n^c} \cdot \bar{f_m^c} / \tau)}},
\end{aligned}
$
}
\label{eq:feat_loss}
\end{equation}
where $\ell_{\text{feat}}^c$ is the partial loss for class $c$, $\tau$ denotes the temperature parameter, and $\1{\cdot}$ denotes the indicator function.

\subsection{Joint Training and Inference}
The overall training objective of our model is as follows.
\begin{equation}
    \mathcal{L}_{\text{total}} = \lambda_{1}\mathcal{L}_{\text{video}} + \lambda_{2}\mathcal{L}_{\text{point}} + \lambda_{3}\mathcal{L}_{\text{score}} + \lambda_{4}\mathcal{L}_{\text{feat}},
\end{equation}
where $\lambda_{*}$ are weighting parameters for balancing the losses, which are determined empirically.

During the test time, we first threshold on the video score $\hat{\mathbf{p}}^{\text{vid}}$ with $\theta^{\text{vid}}$ to determine which action categories are to be localized.
Then, only for the remaining classes, we threshold on the segment-level final scores $\hat{\mathbf{p}}_t$ with $\theta^{\text{seg}}$ to select candidate segments.
Afterward, consecutive candidates are merged into a single proposal, which becomes a localization result.
We set the confidence of each proposal to its outer-inner-contrast score, as in \cite{lee2020background,liu2019completeness}.
To augment the proposal pool, we use multiple thresholds for $\theta^{\text{seg}}$ and perform non-maximum suppression (NMS) to remove overlapping proposals.
Note that the optimal sequence search is not performed at test time, so does not affect the inference time.

\begin{table*}[t]
\centering
\resizebox{.81\textwidth}{!}{
\begin{tabular}{c|l|ccccccc|cc}
\toprule
\multirow{2}{*}{Supervision}       & \multicolumn{1}{c|}{\multirow{2}{*}{Method}} &
\multicolumn{7}{c|}{mAP@IoU (\%)}  & AVG  & AVG  \\
    & \multicolumn{1}{c|}{}  & 0.1  & 0.2  & 0.3  & 0.4  & 0.5  & 0.6  & 0.7 & (0.1:0.5)  & (0.3:0.7)  \\
    \midrule\midrule
\multirow{5}{*}{\begin{tabular}{c}Frame-level\\(Full)\end{tabular}}
    & BMN~\cite{Lin2019BMNBN}  & -  & -  & 56.0  & 47.4  & 38.8  & 29.7  & 20.5  & -  & 38.5 \\
    & P-GCN~\cite{zeng2019p-gcn}  & 69.5  & 67.8  & 63.6  & 57.8  & 49.1  & -  & -  & 61.6  & - \\
    & G-TAD~\cite{xu2020g-tad}  & -  & -  & 54.5  & 47.6  & 40.2  & 30.8  & 23.4  & -  & 39.3 \\
    & BC-GNN~\cite{bai2020bcgnn}  & -  & -  & 57.1  & 49.1  & 40.4  & 31.2  & 23.1  & -  & 40.2 \\
    & Zhao~\etal~\cite{zhao2020bottom-up}  & -  & -  & 53.9  & 50.7  & 45.4  & 38.0  & 28.5  & -  & 43.3 \\
    \midrule
\multirow{5}{*}{\begin{tabular}{c}Video-level\\(Weak)\end{tabular}}
    & Lee~\etal~\cite{lee2021um}  & 67.5  & 61.2  & 52.3  & 43.4  & 33.7  & 22.9  & 12.1  & 51.6  & 32.9  \\
    & CoLA~\cite{Zhang_2021_cola}  & 66.2  & 59.5  & 51.5  & 41.9  & 32.2  & 22.0  & 13.1  & 50.3  & 32.1  \\
    & AUMN~\cite{Luo_2021_AUMN}  & 66.2  & 61.9  & 54.9  & 44.4  & 33.3  & 20.5  & 9.0  & 52.1  & 32.4  \\
    & TS-PCA~\cite{Liu_2021_TS_PCA}  & 67.6  & 61.1  & 53.4  & 43.4  & 34.3  & 24.7  & 13.7  & 52.0  & 33.9  \\
    & UGCT~\cite{Yang_2021_UGCT}  & 69.2  & 62.9  & 55.5  & 46.5  & 35.9  & 23.8  & 11.4  & 54.0  & 34.6  \\
    \midrule
    \multirow{7}{*}{\begin{tabular}{c}Point-level\\(Weak)\end{tabular}}
    & SF-Net$\ssymbol{2}$~\cite{ma2020sfnet}  & 71.0  & 63.4  & 53.2  & 40.7  & 29.3  & 18.4  & 9.6  & 51.5  & 30.2  \\
    & Ju~\textit{et}~\textit{al.}$\ssymbol{2}$~\cite{ju2020point}  & 72.8  & 64.9  & 58.1  & 46.4  & 34.5  & 21.8  & 11.9   & 55.3  & 34.5  \\
    & Ours$\ssymbol{2}$  & \textbf{75.1}  & \textbf{70.5}  & \textbf{63.3}  & \textbf{55.2}  & \textbf{43.9}  & \textbf{33.3}  & \textbf{20.8}  & \textbf{61.6}  &\textbf{43.3}  \\
    \cmidrule(lr){2-11}
    & Moltisanti~\textit{et}~\textit{al.}$\ssymbol{3}$~\cite{Moltisanti2019CVPR} & 24.3  & 19.9  & 15.9  & 12.5  & 9.0  & -  & -  & 16.3  & -  \\
    & SF-Net$\ssymbol{3}$~\cite{ma2020sfnet}  & 68.3  & 62.3  & 52.8  & 42.2  & 30.5  & 20.6  & 12.0   & 51.2  & 31.6  \\
    & Ju~\textit{et}~\textit{al.}$\ssymbol{3}$~\cite{ju2020point}  & 72.3  & 64.7  & 58.2  & 47.1  & 35.9  & 23.0  & 12.8   & 55.6  & 35.4  \\
    & Ours$\ssymbol{3}$  & \textbf{75.7}  & \textbf{71.4}  & \textbf{64.6}  & \textbf{56.5}  & \textbf{45.3}  & \textbf{34.5}  & \textbf{21.8}  & \textbf{62.7}  &  \textbf{44.5}  \\
    \bottomrule
\end{tabular}
}
\caption{
State-of-the-art comparison on THUMOS'14.
We also include the methods under video-level and frame-level supervision for reference.
The average mAPs are computed under the IoU thresholds 0.1:0.5 and 0.3:0.7 with the step size of 0.1. While ${\dagger}$ indicates the use of manually annotated labels from~\cite{ma2020sfnet}, ${\ddagger}$ denotes the use of labels automatically generated in~\cite{Moltisanti2019CVPR}.
}
\label{table:quant_thumos}
\end{table*}

\section{Experiments}
\label{sec:experiments}

\subsection{Experimental Settings}
\noindent\textbf{Datasets.}
THUMOS'14~\cite{THUMOS14} is of 20 action classes with 200 and 213 untrimmed videos for validation and test, respectively.
It is known to be challenging due to the diverse length and the frequent occurrence of action instances.
Following the convention~\cite{nguyen2018weakly}, we use the validation videos for training and test videos for test.
GTEA~\cite{lei2018gtea} contains 28 videos of 7 fine-grained daily actions in the kitchen, among which 21 and 7 videos are utilized for training and test, respectively.
BEOID~\cite{damen2014BEOID} has 58 videos with a total of 30 action categories.
We follow the data split provided by \cite{ma2020sfnet}.
ActivityNet~\cite{caba2015activitynet} is a large-scale dataset with two versions.
The version 1.3 includes 10,024 training, 4,926 validation, and 5,044 test videos with 200 action classes.
The version 1.2 consists of 4,819 training, 2,383 validation, and 2,480 test videos with 100 categories.
We evaluate our model on the validation sets for both versions.
It should be noted that our model takes only point-level annotations for training.

\noindent\textbf{Evaluation metrics.}
Following the standard protocol of temporal action localization, we compute mean average precisions (mAPs) under several different levels of intersection over union (IoU) thresholds.
We note that performances at small IoU thresholds demonstrate the ability in finding actions, while those under high IoU thresholds exhibit the completeness of action predictions.

\noindent\textbf{Implementation details.}
We employ the two-stream I3D networks~\cite{carreira2017quo} pre-trained on Kinetics-400~\cite{carreira2017quo} as our feature extractor, which is not fine-tuned in our experiments for fair comparison.
To obtain optical flow maps, we use TV-L1 algorithm~\cite{wedel2009improved}.
Each video is split into 16-frame segments, which are taken as inputs by the feature extractor resulting in 1024-dim features for each modality (\ie, $D=2048$).
We use the original number of segments as $T$ without sampling.
Our model is optimized by Adam~\cite{kingma2014adam} with the learning rate of $10^{-4}$ and the batch size of 16.
Hyper-parameters are determined by grid search: $\gamma=0.95$, $\tau=0.1$.
The video-level threshold $\theta^{\text{vid}}$ is set to 0.5, while the segment-level threshold $\theta^{\text{seg}}$ spans from 0 to 0.25 with a step size of 0.05.
The NMS is performed with the threshold of 0.6.

\subsection{Comparison with State-of-the-art Methods}
In \Tref{table:quant_thumos}, we compare our method with state-of-the-art models under different levels of supervision on THUMOS'14.
We note that fully-supervised models require far more expensive annotation costs compared to weakly-supervised counterparts.
In the comparison, our model significantly outperforms the state-of-the-art point-supervised approaches.
We also notice the large performance margins at high IoU thresholds, \eg, $\sim$11\% in mAP@0.6 and $\sim$9\% in mAP@0.7.
This confirms that the proposed method aids in locating the complete action instances.
At the same time, our model largely surpasses the video-supervised methods with the comparable labeling cost.
Further, our model even performs favorably against the fully-supervised methods in terms of average mAPs at the much lower annotation cost.
It is, however, also shown that ours lags behind them at high IoU thresholds, due to the lack of boundary information.

We provide the experimental results on GTEA and BEOID benchmarks in \Tref{table:quant_gtea_beoid}.
On the both datasets, our method beats the existing state-of-the-art methods with a large gap.
Notably, our method shows significant performance boosts under the high thresholds of 0.5 and 0.7, verifying the efficacy of the proposed completeness learning.

\Tref{table:quant_anet12} and \Tref{table:quant_anet13} summarize the results on ActivityNet.
Our model shows the superior performances over all the existing weakly-supervised approaches on both versions.
It can be also observed that the performance gains upon video-level labels are relatively small compared to THUMOS'14, which we conjecture is due to the far less frequent action instances (1.5 \textit{vs.} 15 instances per video).

\begin{table}[!t]
\centering
\resizebox{0.86\textwidth}{!}{
\begin{tabular}{c|l|cccc|c}
\toprule
\multirow{2}{*}{Dataset}       & \multicolumn{1}{c|}{\multirow{2}{*}{Method}} &
\multicolumn{4}{c|}{mAP@IoU (\%)}  & \multirow{2}{*}{AVG}  \\
    & \multicolumn{1}{c|}{}  & 0.1  & 0.3  & 0.5  & 0.7 &  \\
    \midrule\midrule
\multirow{5}{*}{\begin{tabular}{c}GTEA\end{tabular}}
    & SF-Net~\cite{ma2020sfnet}  & 58.0  & 37.9  & 19.3  & 11.9  & 31.0  \\
    & SF-Net$\ssymbol{1}$~\cite{ma2020sfnet}  & 52.9  & 37.6  & 21.7  & 13.7  & 31.1  \\
    & Ju~\textit{et}~\textit{al.}~\cite{ju2020point}  & 59.7  & 38.3  & 21.9  & 18.1  & 33.7  \\
    & Li~\textit{et}~\textit{al.}~\cite{li2021seg-timestamp}  & 60.2  & 44.7  & 28.8  & 12.2  & 36.4  \\
    & Ours  & \textbf{63.9}  & \textbf{55.7}  & \textbf{33.9}  & \textbf{20.8}  & \textbf{43.5}  \\
    \midrule
\multirow{5}{*}{\begin{tabular}{c}BEOID\end{tabular}}
    & SF-Net~\cite{ma2020sfnet}  & 62.9  & 40.6  & 16.7  & 3.5  & 30.9  \\
    & SF-Net$\ssymbol{1}$~\cite{ma2020sfnet}  & 64.6  & 42.2  & 27.3  & 12.2  & 36.5  \\
    & Ju~\textit{et}~\textit{al.}~\cite{ju2020point}  & 63.2  & 46.8  & 20.9  & 5.8  & 34.9  \\
    & Li~\textit{et}~\textit{al.}~\cite{li2021seg-timestamp}  & 71.5  & 40.3  & 20.3  & 5.5  & 34.4  \\
    & Ours  & \textbf{76.9}  & \textbf{61.4}  & \textbf{42.7}  & \textbf{25.1}  & \textbf{51.8}  \\
    \bottomrule
\end{tabular}
}
\caption{
State-of-the-art comparison on GTEA and BEOID. AVG denotes the average mAP at the thresholds 0.1:0.1:0.7.
* denotes the reproduced results by official implementation.
}
\label{table:quant_gtea_beoid}
\end{table}

\begin{table}[t]
\centering
\resizebox{0.8\columnwidth}{!}{
\begin{tabular}{c|l|ccc|c}
\toprule
\multirow{2}{*}{Supervision} &
\multicolumn{1}{c|}{\multirow{2}{*}{Method}} &
\multicolumn{3}{c|}{mAP@IoU (\%)} & \multirow{2}{*}{AVG} \\
        &  & 0.5  & 0.75  & 0.95  &  \\
      \midrule\midrule
\multirow{1}{*}{Frame-level}
        & SSN~\cite{zhao2017temporal}  & 41.3  & 27.0  & 6.1  & 26.6    \\
        \midrule
\multirow{4}{*}{Video-level} 
       & Lee~\etal~\cite{lee2021um}  & 41.2  & 25.6  & 6.0  & 25.9  \\
       & AUMN~\cite{Luo_2021_AUMN}  & 42.0  & 25.0  & 5.6  & 25.5  \\
       & UGCT~\cite{Yang_2021_UGCT}  & 41.8  & 25.3  & 5.9  & 25.8  \\
       & CoLA~\cite{Zhang_2021_cola}  & 42.7  & 25.7  & 5.8  & 26.1  \\
       \midrule
\multirow{2}{*}{Point-level} 
       & SF-Net~\cite{ma2020sfnet}  & 37.8  & -  & -  & 22.8  \\
       & Ours  & \textbf{44.0}  & \textbf{26.0}  & \textbf{5.9}  & \textbf{26.8}  \\
    \bottomrule
\end{tabular}
}
\caption{
State-of-the-art comparison on ActivityNet~1.2. AVG is the averaged mAP at the thresholds 0.5:0.05:0.95.
}
\label{table:quant_anet12}
\end{table}

\subsection{Analysis}
\label{subsec:analysis}

\noindent\textbf{Effect of each component.}
In \Tref{table:ablation_thumos_v2}, we conduct ablation study to investigate the contribution of each component.
The upper section reports the baseline performances, from which we observe a large score gain brought by the point-level supervision, especially under low IoU thresholds.
It mainly comes from the background modeling~\cite{lee2020background,lee2021um,Nguyen2019WeaklySupervisedAL} and the help of point annotations in spotting action instances.
On the other hand, the lower section demonstrates the results of the proposed method, where completeness guidance is provided for the model.
We observe the absolute average mAP gains of 4.7\% and 1.7\% from the proposed contrastive losses regarding score and feature similarity, respectively.
Moreover, with the two losses combined, the performance is further boosted to 52.8\%.
This clearly shows that the proposed two losses are complementary and beneficial for precise action localization.
Notably, the scores at high IoU thresholds are largely improved, verifying the efficacy of our completeness learning.

\begin{table}[t]
\centering
\resizebox{.8\columnwidth}{!}{
\begin{tabular}{c|l|ccc|c}
\toprule
\multirow{2}{*}{Supervision} &
\multicolumn{1}{c|}{\multirow{2}{*}{Method}} &
\multicolumn{3}{c|}{mAP@IoU (\%)} & \multirow{2}{*}{AVG} \\
        &  & 0.5  & 0.75  & 0.95  &  \\
        \midrule\midrule
\multirow{5}{*}{Frame-level}
        & BMN~\cite{Lin2019BMNBN}  & 50.1  & 34.8  & 8.3  & 33.9  \\
        & P-GCN~\cite{zeng2019p-gcn}  & 48.3  & 33.2  & 3.3  & 31.1    \\
        & G-TAD~\cite{xu2020g-tad}  & 50.4  & 34.6  & 9.0  & 34.1    \\
        & BC-GNN~\cite{bai2020bcgnn}  & 50.6  & 34.8  & 9.4  & 34.2    \\
        & Zhao~\etal~\cite{zhao2020bottom-up}  & 43.5  & 33.9  & 9.2  & 30.1    \\
        \midrule
\multirow{3}{*}{Video-level} 
       & Lee~\etal~\cite{lee2021um}  & 37.0  & 23.9  & 5.7  & 23.7  \\
       & AUMN~\cite{Luo_2021_AUMN}  & 38.3  & 23.5  & 5.2  & 23.5  \\
       & TS-PCA~\cite{Yang_2021_UGCT}  & 37.4  & 23.5  & 5.9  & 23.7  \\
       \midrule
\multirow{1}{*}{Point-level}
       & Ours  & \textbf{40.4}  & \textbf{24.6}  & \textbf{5.7}  & \textbf{25.1}  \\
\bottomrule
\end{tabular}
}
\caption{
State-of-the-art comparison on ActivityNet~1.3. AVG is the averaged mAP at the thresholds 0.5:0.05:0.95.
}
\label{table:quant_anet13}
\end{table}
\begin{table}[t]
\centering
\resizebox{0.89\textwidth}{!}{
\begin{tabular}{cccc|cccc|c}
    \toprule
    \multirow{2}{*}{$\mathcal{L_{\text{video}}}$} & \multirow{2}{*}{$\mathcal{L_{\text{point}}}$} & \multirow{2}{*}{$\mathcal{L_{\text{score}}}$} & \multirow{2}{*}{$\mathcal{L_{\text{feat}}}$}
    & \multicolumn{4}{c|}{mAP@IoU (\%)}  & \multirow{2}{*}{AVG}  \\
      &  &  &  & 0.1  & 0.3  & 0.5  & 0.7  &   \\
    \midrule\midrule
    \ForestGreencheck  & \redx & \redx & \redx & 51.9  & 37.1  & 20.3  & 6.0  & 28.7  \\
    \ForestGreencheck  & \ForestGreencheck  & \redx & \redx & 70.7  & 58.1  & 40.7  & 16.1  & 47.3  \\
    \midrule
    \ForestGreencheck  & \ForestGreencheck  & \ForestGreencheck & \redx  & 75.1  & 64.4  & 44.5  & 20.0  & 52.0  \\
    \ForestGreencheck  & \ForestGreencheck  & \redx  & \ForestGreencheck & 72.1  & 60.5  & 42.1  & 17.9  & 49.0  \\
    \ForestGreencheck  & \ForestGreencheck  & \ForestGreencheck  & \ForestGreencheck  & \textbf{75.7}  & \textbf{64.6}  & \textbf{45.3}  & \textbf{21.8}  & \textbf{52.8}  \\
    \bottomrule
\end{tabular}
}
\caption{
Ablation study on THUMOS'14. AVG represents the average mAP at the IoU thresholds 0.1:0.1:0.7.
}
\label{table:ablation_thumos_v2}
\end{table}
\begin{table}[t]
\centering
\resizebox{0.92\textwidth}{!}{
\begin{tabular}{c|c|cccc|c}
\toprule
\multicolumn{1}{c|}{\multirow{2}{*}{Scoring method}} & \multirow{2}{*}{\begin{tabular}{c}Sequence\\accuracy\end{tabular}}       &
\multicolumn{4}{c|}{mAP@IoU (\%)}  & \multirow{2}{*}{AVG}  \\
    \multicolumn{1}{c|}{}  &  & 0.1 & 0.3  & 0.5  & 0.7  &  \\
    \midrule\midrule
    Baseline  & N/A  & 70.7  & 58.1  & 40.7  & 16.1  & 47.3  \\
    \midrule
    (a) Inner scores  & 74.0  & 74.7  & 61.4  & 40.9  & 15.2  & 49.0  \\
    (b) Contrast-act  & 80.1  & 74.3  & 63.3  & 43.6  & 19.5  & 50.8  \\
    (c) Contrast-both  & \textbf{83.9}  & \textbf{75.7}  & \textbf{64.6}  & \textbf{45.3}  & \textbf{21.8}  & \textbf{52.8}  \\
    \bottomrule
\end{tabular}
}
\caption{
Comparison of different scoring methods for optimal sequence search on THUMOS'14.
AVG denotes the average mAP at the IoU thresholds 0.1:0.1:0.7.
}
\label{table:ablation_scoring}
\end{table}

\begin{figure*}[t]
  \centering
  \includegraphics[clip=true, width=0.94\textwidth]{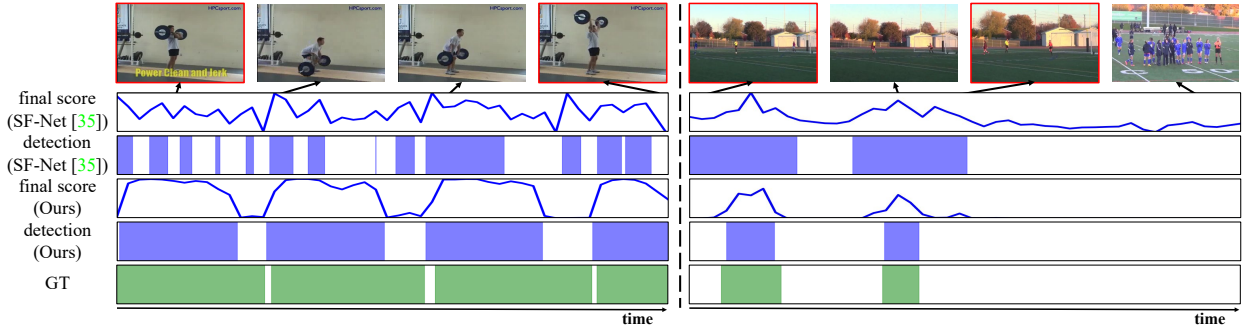}
  \caption{Qualitative comparison with SF-Net~\cite{ma2020sfnet} on THUMOS'14.
  We provide two examples with different action classes: (1) \textit{CleanAndJerk} and (2) \textit{SoccerPenalty}.
  For each video, we present final scores and detection results from SF-Net and our model as well as ground truth action interval.
  The detection threshold is set to 0.2 for our method and set to the mean score for SF-Net following the original paper.
  The red boxes indicate the frames that are misclassified by SF-Net but detected by our method.
  Note that all of our detection results show high IoUs ($>$ 0.6) with the ground-truths.
  }
  \label{fig:qualitative_comparison}
\end{figure*}
\noindent\textbf{Comparison of different scoring methods.}
In \Tref{table:ablation_scoring}, we compare different sequence scoring methods regarding frame-level accuracy of optimal sequences in the training set as well as localization performances in the test set of THUMOS'14.
Specifically, we investigate three variants: (a) inner scores and (b) score contrast of action instances, and (c) contrast of both action and background ones.
As a result, compared to inner scores, the contrast methods generate more accurate optimal sequences and bring larger performance gains at high IoU thresholds.
Moreover, we observe that incorporating background instances for score calculation helps to find highly accurate optimal sequences, thereby improving the localization performance at test time.

\begin{table}[t]
\centering
\resizebox{0.96\textwidth}{!}{
\begin{tabular}{c|c|c|ccc|c}
\toprule
\multirow{2}{*}{Method}       & \multicolumn{1}{c|}{\multirow{2}{*}{Distribution}} & \multirow{2}{*}{\begin{tabular}{c}Sequence\\accuracy\end{tabular}}       &
\multicolumn{3}{c|}{mAP@IoU (\%)}  & \multirow{2}{*}{AVG}  \\
    & \multicolumn{1}{c|}{}  & & 0.3  & 0.5  & 0.7  &  \\
    \midrule\midrule
\multirow{3}{*}{\begin{tabular}{c}SF-Net~\cite{ma2020sfnet}\end{tabular}}
    & Manual  & N/A  & 53.3  & 28.8  & 9.7  & 40.6  \\
    & Uniform  & N/A  & 52.0  & 30.2  & 11.8  & 40.5  \\
    & Gaussian  & N/A  & 47.4  & 26.2  & 9.1  & 36.7  \\
    \midrule
\multirow{3}{*}{\begin{tabular}{c}Ju~\etal~\cite{ju2020point}\end{tabular}}
    & Manual  & N/A  & 58.1  & 34.5  & 11.9  & 44.3  \\
    & Uniform  & N/A  & 55.6  & 32.3  & 12.3  & 42.9  \\
    & Gaussian  & N/A  & 58.2  & 35.9  & 12.8  & 44.8  \\
    \midrule
\multirow{3}{*}{\begin{tabular}{c}Ours\end{tabular}}
    & Manual  & 83.7  & 63.3  & 43.9  & 20.8  & 51.7  \\
    & Uniform  & 76.6  & 60.4  & 42.6  & 20.2  & 49.3 \\
    & Gaussian  & \textbf{83.9}  & \textbf{64.6}  & \textbf{45.3}  & \textbf{21.8}  & \textbf{52.8}  \\
    \bottomrule
\end{tabular}
}
\caption{
Comparison of the point-level labels from different distributions on THUMOS'14.
AVG denotes the average mAP at the IoU thresholds 0.1:0.1:0.7.
}
\label{table:ablation_label}
\end{table}

\noindent\textbf{Comparison of different label distributions.}
In \Tref{table:ablation_label}, we explore different label distributions.
``Manual'' indicates the use of human annotations from \cite{ma2020sfnet}, whereas the others denote the simulated labels from the corresponding distributions.
It is shown that our method significantly outperforms the existing methods regardless of the distribution choice, showing its robustness.
We also observe that our method performs slightly worse in ``Uniform'' compared to the other distributions.
We conjecture this is because less discriminative points have more chances to be annotated.
Their neighbors are likely to have lower confidence, probably leading to sub-optimal sequences by the greedy algorithm.
Indeed, the optimal sequence accuracy is shown to be the lowest in the uniform distribution, which supports our claim.

\subsection{Qualitative Comparison}

We present qualitative comparisons with SF-Net~\cite{ma2020sfnet} in \Fref{fig:qualitative_comparison}.
It can be clearly noticed that our method locates the action instances more precisely.
Specifically, in the left example, SF-Net produces fragmentary predictions with false negatives, whereas our method detects the complete action instances without splitting them.
In the right sample, while SF-Net overestimates the action instances with false positives, our method produces precise detection results by contrasting action frames from background ones well.
The red boxes highlight the false negatives and false positives of SF-Net in the left and right examples, respectively.
We note that all the predictions of our model in both examples have high IoUs larger than 0.6 with the corresponding ground-truth instances, validating the effectiveness of our completeness learning.
Comparisons on other benchmarks and more visualization results can be found in Sec.~\textcolor{red}{C} of the appendix.

\section{Conclusion}
\label{sec:conclusion}

In this paper, we presented a new framework for point-supervised temporal action localization, where dense sequences provide completeness guidance to the model.
Concretely, we find the optimal sequence consistent with point labels based on the completeness score, which is efficiently implemented with a greedy algorithm.
To learn completeness from the obtained sequence, we introduced two novel losses which encourage contrast between action and background instances regarding action score and feature similarity, respectively.
Experiments validated that the optimal sequences are accurate and the proposed losses indeed help to detect complete action instances.
Moreover, our model achieves a new state-of-the-art with a large gap on four benchmarks.
Notably, it even outperforms fully-supervised methods on average despite the lower supervision level.

\section*{Acknowledgements}
{\small
\noindent This project was partly supported by the National Research Foundation of Korea grant funded by the Korea government (MSIT) (No. 2019R1A2C2003760) and the Institute for Information \& Communications Technology Planning \& Evaluation (IITP) grant funded by the Korea government (No. 2020-0-01361: Artificial Intelligence Graduate School Program (YONSEI UNIVERSITY)).
}

\setcounter{section}{0}
\renewcommand\thesection{\Alph{section}}

\section{Regarding Point-level Supervision}
\label{sec:supp_point}

In this paper, we tackle temporal action localization under point-level supervision.
Here, timestamp are denoted by ``points" in the temporal axis, whereas ``points" have also been widely used to represent spatial pixels in the literature.
Bearman~\etal~\cite{Bearman2016whats_the_points} introduce the first weakly-supervised semantic segmentation framework that takes as supervision a single annotated pixel for each object.
Since that work, a great amount of efforts~\cite{ke2021universal_segmentation,laradji2021point_covid,laradji2020point_instance2,ren2020ufo,zhou2019point_instance} have been endeavored to utilize point-level supervision to solve various segmentation tasks in images or videos, thanks to its affordable annotation cost.
Meanwhile, there are also attempts to employ point-level supervision to train object detectors~\cite{mcever2020pcams,papadopoulos2017extreme_detection,papadopoulos2017click_detection}.
On the other hand, spatial points have also been explored to provide supervision for the weakly-supervised spatio-temporal action localization task~\cite{mettes2019point_action,mettes2016spot_on}.

We remark that the definition of ``point" in our problem setting is based on the temporal dimension, differing from that of the work above.

\section{Greedy Optimal Sequence Search}
\label{sec:supp_algorithm}

As discussed in the main paper, the search space of optimal sequence selection would grow exponentially as the length of the input video increases, which makes the optimal sequence search intractable.
To bypass the cost issue, we design a greedy algorithm that makes locally optimal choices at each step under a fixed budget.
Specifically, we process an input video in a sequential way, taking one segment at a timestep.
At each timestep $t$, we consider all possible $t$-length candidate sequences consistent with point labels, and compute their completeness scores by averaging contrast scores of the action and background instances constituting the sequences (Eq.~\eqref{eq:score} of the main paper).
In this calculation, we do not include the ongoing (\ie, not terminated) instance, as it is infeasible to derive its contrast score without looking ahead to the future.
Afterwards, we keep only the top $\alpha$ (budget size) candidates regarding the completeness scores.
When the step $t$ reaches the end of the video, we terminate the algorithm and select the optimal sequence with the highest score.
In this way, we can save a large amount of the computational cost, thereby making the search process tractable.
The pseudo-code of our algorithm for class $c$ is described in \Aref{alg:viterbi}.

Since the budget $\alpha$ affects the computational cost as well as the performance, we investigate several different budget sizes on THUMOS'14.
For the computational cost, we train the model for 100 epochs and report the average execution time of optimal sequence selection for an epoch (\ie, 200 training videos).
The selection is implemented in multiprocessing with 16 worker processes and performed on a single AMD-3960X Threadripper CPU.
\Tref{table:ablation_budget} shows the average mAPs (\%) and the execution times (sec) with varying $\alpha$.
As can be expected, when the budget increases, the computational cost grows in a nearly linear way.
Besides, when $\alpha$ is set to a too-small value (\eg, 1), the selected optimal sequence is likely to be a local optimum, leading to a significant performance drop.
On the other hand, the performance differences are insignificant when $\alpha$ is larger than 5.
This indicates that the model is fairly robust against the budget size and a not-too-small $\alpha$ is sufficient to find the sequences that can provide helpful completeness guidance to the model.
In practice, we set $\alpha$ to 25, as it achieves the best performance at an affordable cost of fewer than 5 seconds for processing the whole training videos.

\begin{table}[t]
\caption{
Analysis on the budget size $\alpha$ on THUMOS'14. We provide the execution times as well as the average mAPs under IoU thresholds 0.1:0.1:0.7 with varying $\alpha$ from 1 to 100.
The average execution time for optimal sequence selection per epoch is reported in seconds.
}
\centering
\resizebox{1.0\textwidth}{!}{
\begin{tabular}{c|cccccc}
\toprule
\multicolumn{1}{r|}{$\alpha$~$\longrightarrow$} & 1 & 5 & 10 & 25 & 50 & 100 \\
\midrule\midrule
mAP@AVG (\%) & 51.3 & 52.5 & 52.6 & 52.8  & 52.7  & 52.7  \\
Execution time (sec) & 0.683 & 1.343 & 2.151 & 4.398  & 8.512  & 16.769  \\
\bottomrule
\end{tabular}
}
\label{table:ablation_budget}
\end{table}
\begin{figure}[t]
  \centering
  \includegraphics[clip=true, width=0.93\textwidth]{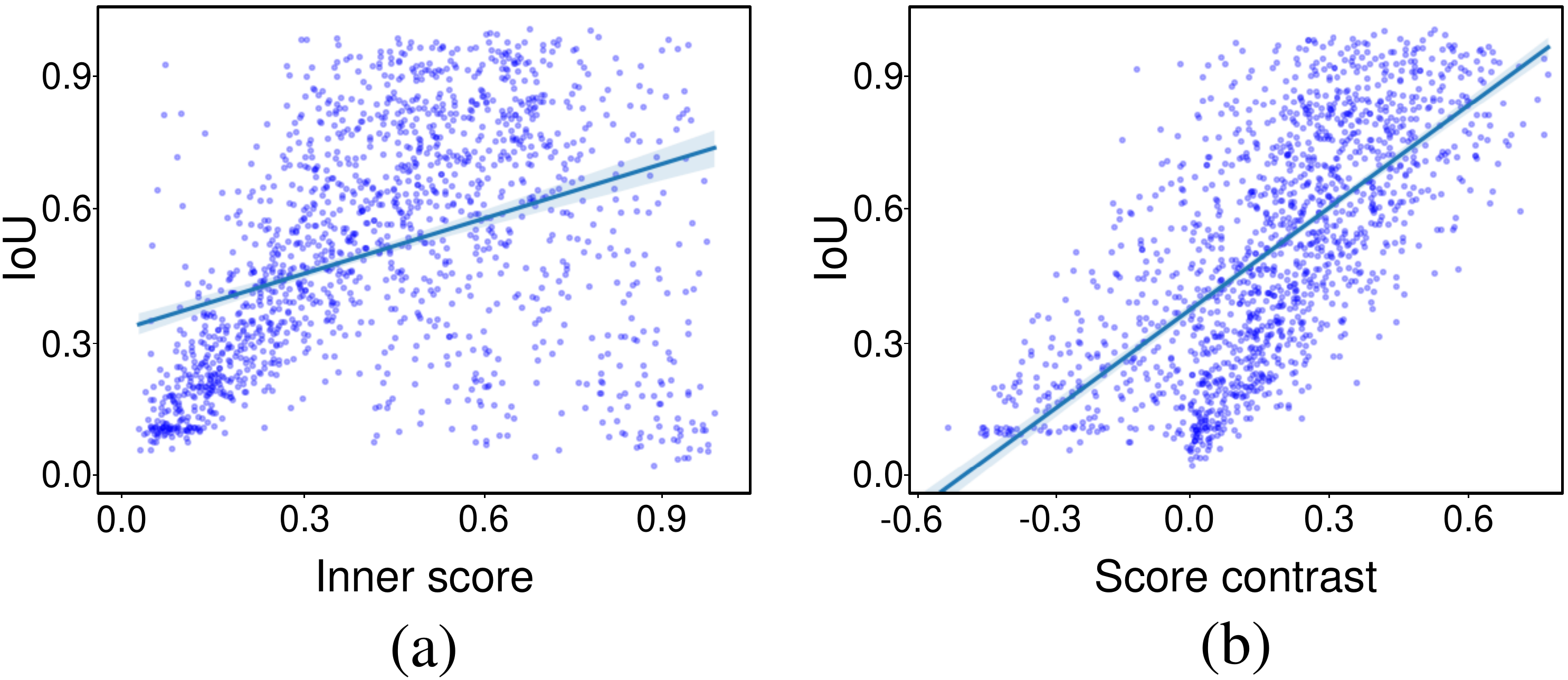}
  \caption{Correlation between scores and IoUs with ground-truths. (a) The inner score shows moderate correlation (Pearson’s r = 0.38), whereas (b) the score contrast displays much stronger correlation (Pearson's r = 0.68).}
  \label{fig:correlation}
\end{figure}

\section{Additional Experiments}
\label{sec:supp_experiments}

\subsection{Score contrast \textit{vs.} completeness}
To analyze the correlation between score contrast and action completeness, we draw the scatter plot of score contrast \textit{vs.} IoUs with ground-truth action instances, using the randomly sampled 2,000 temporal intervals in the THUMOS'14 training videos.
For reference, we also present the scatter plot of inner action scores \textit{vs.} IoUs with the same intervals.
In the experiments, we use the \textit{baseline} model for fair comparison.
\Fref{fig:correlation}a demonstrates that there is a moderate correlation between inner action scores and IoUs, but there are many cases with large inner scores but low IoUs (see bottom right).
On the contrary, as shown in \Fref{fig:correlation}b, score contrast correlates much stronger with IoUs, demonstrating its efficacy as a proxy for measuring the action completeness without any supervision.

\clearpage

\begin{algorithm*}[t]
  \caption{Greedy Optimal Sequence Search}
  \label{alg:viterbi}
\begin{algorithmic}[1]
  \REQUIRE \parbox[t]{0.99\columnwidth}{
  class-specific action points (ascending) $\mathcal{B}^\text{act}_c=\{t_{i}^\text{act}\}_{i=1}^{M^\text{act}_c}$, pseudo background points (ascending) $\mathcal{B}^\text{bkg}=\{t_{j}^\text{bkg}\}_{j=1}^{M^\text{bkg}}$,\\
  the number of class-specific action points $M_c^{\text{act}}$, the number of pseudo background points $M^{\text{bkg}}$,\\
  fixed budget size $\alpha$}
  \ENSURE optimal sequence $\pi_{c}^{*}$
  \COMMENT{\\~\\//~~Definition: $\pi_c=\{(s_n, e_n, z_n)\}_{n=1}^{N}, \mathcal{S}_c=\{(\pi_c, \mathcal{R}(\pi_c))\}$ (refer to Sec. \textcolor{red}{3.2} of the main paper for the definition of $\pi_c$ and $\mathcal{R}(\pi_c)$)}
  \COMMENT{\\//~~Initialize the first instance $(s_1=e_1=1)$ with the same category as that of the first point label\\[0.1em]}
  \STATE \textbf{if} $t_{1}^{\text{act}} > t_{1}^{\text{bkg}}$,~~\textbf{then} $\pi_c^0 \gets \{(1,1,0)\}$~~\textbf{else} $\pi_c^0 \gets \{(1,1,1)\}$
  \STATE $\mathcal{S}_c \gets \{(\pi_c^0, \infty)\}$
  \STATE $i \gets 1;~~j \gets 1$
  \COMMENT{\\~\\//~~For each step $t$, find the top $\alpha$ sequences which span from the first segment to the $t$-th segment while agreeing with point labels. \\[0.1em]}
  \FOR{$t = 2$ {\bfseries to} $T$}
  \STATE{\bgroup\small\textsl{//~~Find the upcoming points for action and background, respectively.}\egroup}
  \STATE \textbf{if} $t > t_{i}^{\text{act}}$,~~\textbf{then} $i \gets \min{(i+1,M_c^{\text{act}})}$;~~\textbf{if} $t > t_{j}^{\text{bkg}}$,~~\textbf{then} $j \gets \min{(j+1,M^{\text{bkg}})}$ 
  \COMMENT{\\//~~Remember the category of the closest upcoming point, as it will determine the possible cases (to continue or to be terminated)\\[0.1em]}
  \STATE \textbf{if} $t_{i}^{\text{act}} > t_{j}^{\text{bkg}}$,~~\textbf{then} $z^{\text{upcoming}} \gets 0$~~\textbf{else} $z^{\text{upcoming}} \gets 1$
  \COMMENT{\\//~~If $t$ surpasses either of the last points for action and background, reverse the upcoming category\\[0.1em]}
  \STATE \textbf{if} $t > \min{(t_{i}^{\text{act}}, t_{j}^{\text{bkg}})}$,~~\textbf{then} $z^{\text{upcoming}} \gets 1 - z^{\text{upcoming}}$
  \COMMENT{\\~\\//~~Update the candidate sequence set for the timestep $t$\\[0.1em]}
  \STATE $\mathcal{S}_c^{\text{next}} \gets \varnothing$
  \WHILE{$\mathcal{S}_c\neq \varnothing$}
  \STATE pop $\big(\pi_c=\{(s_n, e_n, z_n)\}_{n=1}^{N}$, $\mathcal{R}^{\text{current}}\big)$ from $\mathcal{S}_c$
  \STATE pop the last instance $(s_N, e_N, z_N)$ from $\pi_c$ ~~~~~~~~~~~~~~~~~//~~\bgroup\small\textsl{$e_N$ should be equal to $t-1$}\egroup
  \COMMENT{\\//~~The case where the last instance continues at timestep $t$\\[0.1em]}
  \IF{$z_N=z^{\text{upcoming}}$ \textbf{or} $t \not\in \big(\mathcal{B}_c^{\text{act}} \cup \mathcal{B}^{\text{bkg}}\big)$}
  \STATE $\pi_c^{\text{new}} \gets \pi_c \cup \{(s_N, e_N+1, z_N)\}$
  \STATE $\mathcal{S}_c^{\text{next}} \gets \mathcal{S}_c^{\text{next}} \cup \{\big(\pi_c^{\text{new}}, \mathcal{R}^{\text{current}}\big)\}$
  \ENDIF
  \COMMENT{\\//~~The case where the last instance is terminated at timestep $t-1$ and a new instance starts at timestep $t$\\[0.1em]}
  \IF{$z_N \neq z^{\text{upcoming}}$}
  \STATE $\pi_c^{\text{last}} \gets \{(s_N, e_N, z_N)\}$
  \COMMENT{\\//~~Update the score of the candidate sequence by averaging the contrast scores again\\[0.1em]}
  \STATE \textbf{if} $N=1$,~~\textbf{then} $\mathcal{R}^{\text{new}} \gets \mathcal{R}(\pi_c^{\text{last}})$~~\textbf{else} $\mathcal{R}^{\text{new}} \gets \big(\mathcal{R}(\pi_c^{\text{last}}) + (N-1)\mathcal{R}^{\text{current}}\big)/N$
  \COMMENT{\\//~~Create a new instance that starts right after the last instance, with the category of $z^{\text{upcoming}}$\\[0.1em]}
  \STATE $\pi_c^{\text{new}} \gets \pi_c \cup \pi_c^{\text{last}} \cup \{(e_{N}+1, e_{N}+1, z^{\text{upcoming}})\}$
  \STATE $\mathcal{S}_c^{\text{next}} \gets \mathcal{S}_c^{\text{next}} \cup \{\big(\pi_c^{\text{new}}, \mathcal{R}^{\text{new}}\big)\}$
  \ENDIF
  \ENDWHILE
  \STATE $\mathcal{S}_c \gets \mathcal{S}_c^{\text{next}}$
  \COMMENT{\\//~~Pruning with the budget size $\alpha$\\[0.1em]}
  \WHILE{$|\mathcal{S}_c| > \alpha$}
  \STATE $\pi_c^{\text{min}} \gets \argmin_{\pi_c}\mathcal{R}^{\text{current}}$ for $\big(\pi_c, \mathcal{R}^{\text{current}}\big) \in \mathcal{S}_c$
  \STATE pop $\big(\pi_c^{\text{min}}, \mathcal{R}^{\text{current}}\big)$ from $\mathcal{S}_c$
  \ENDWHILE
  \ENDFOR
  \COMMENT{\\~\\//~~Return the optimal sequence\\[0.1em]}
  \STATE $\pi_c^* \gets \argmax_{\pi_c}\mathcal{R}(\pi_c)$ for $\pi_c \in \mathcal{S}_c$
  \STATE \textbf{return} $\pi_c^*$
\end{algorithmic}
\end{algorithm*}
\clearpage

\begin{table}[t]
\caption{
Comparison of different pseudo background mining approaches on THUMOS'14. AVG represents the average mAP at the IoU thresholds 0.1:0.1:0.7.
}
\centering
\resizebox{1.0\textwidth}{!}{
\begin{tabular}{c|ccccccc|c}
\toprule
\multirow{2}{*}{Mining approach} 
& \multicolumn{8}{c}{mAP@IoU (\%)}  \\
& 0.1  & 0.2  & 0.3  & 0.4  & 0.5  & 0.6  & 0.7  & AVG  \\
\midrule\midrule
Global mining~\cite{ma2020sfnet}  & 67.4  & 61.1  & 54.9  & 46.3  & 36.4  & 25.7  & 13.4  & 43.6  \\
Ours w/o filling  & 70.1  & 64.4  & 57.6  & 49.5  & 39.4  & 29.5  & 15.5  & 46.6  \\
Ours  & 70.7  & 65.2  & 58.1  & 49.8  & 40.7  & 30.2  & 16.1  & 47.3  \\
\bottomrule
\end{tabular}
}
\label{table:ablation_bkg_mining}
\end{table}

\subsection{Analysis on Pseudo Background Mining}
We compare different variants of pseudo background mining on THUMOS'14.
Specifically, we consider three variants: (1) ``Global mining'' selects the top $\eta M^{\text{act}}$ points throughout the whole video without considering their locations as in SF-Net~\cite{ma2020sfnet}, where $M^{\text{act}}$ is the number of action instances and $\eta$ is set to 5,
(2) ``Ours w/o filling'' follows the principle described in \Sref{subsec:baseline} except the filling stage, \ie, we select at least one background point for each section between two action points, and (3) ``Ours'' mines all points between the background points for each section if multiple points are found in the second variant. Note that we use the \textit{baseline} model without completeness learning for clear comparison.

The results are demonstrated in \Tref{table:ablation_bkg_mining}.
It can be observed that both of our methods significantly outperform the ``Global mining'' approach, which verifies the effectiveness of our selection principle that at least one background point should be placed for each section.
Moreover, by ensuring at least one background point for each section, the search space of optimal sequence selection can be significantly reduced, although we do not include the cost analysis for this experiment.
Meanwhile, we notice that filling between two background points slightly boosts the localization performance.
This is presumably because hard background points with low background scores can be collected in the filling step.

\subsection{Optimal Sequence Visualization}
\begin{figure*}[t]
  \centering
  \includegraphics[clip=true, width=0.99\textwidth]{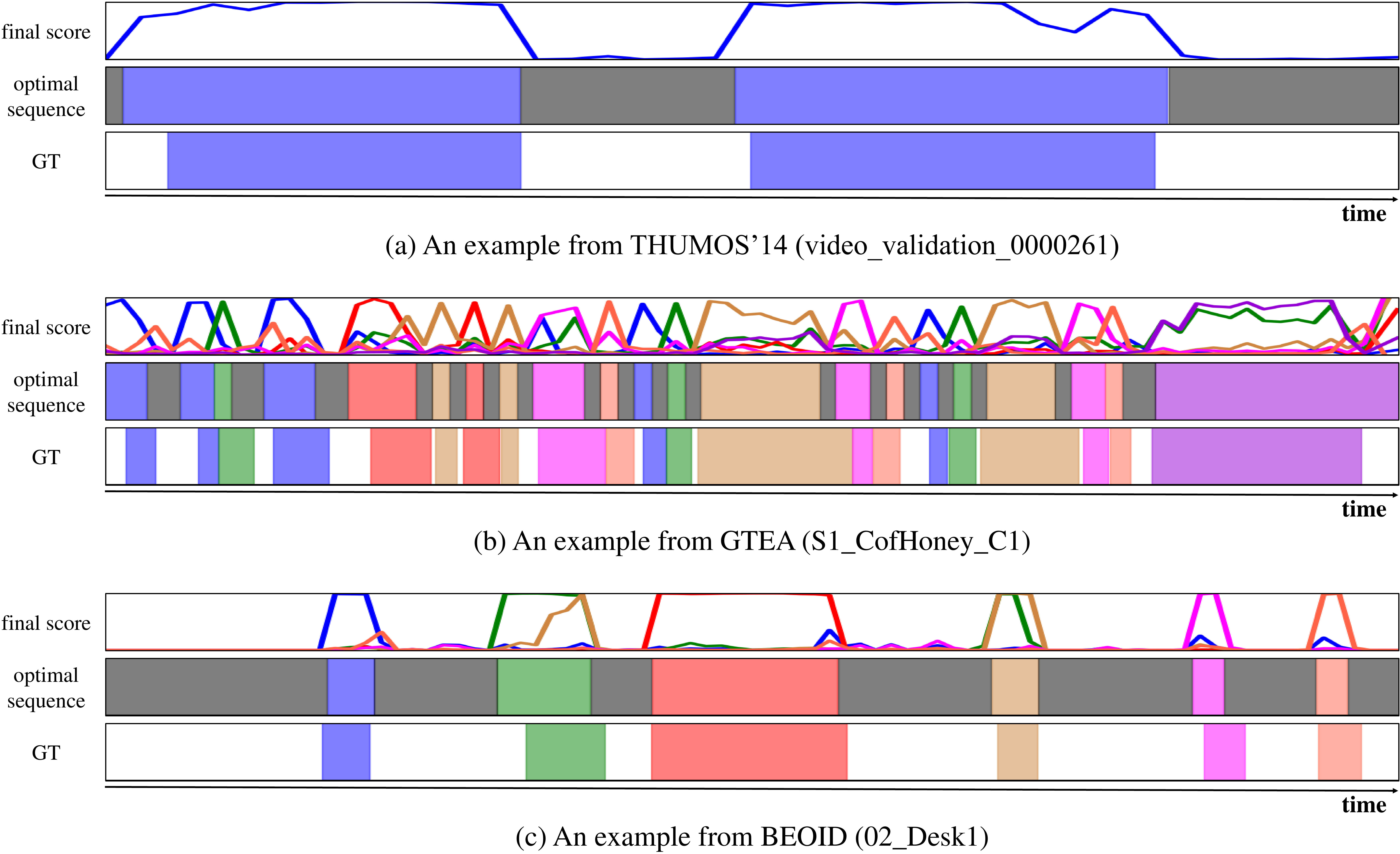}
  \caption{Optimal sequence visualization on the three benchmarks.
  The examples are taken from (a) THUMOS'14, (b) GTEA, and (c) BEOID, respectively.
  Note that all of the examples belong to the training set of the corresponding benchmarks.
  For each video, we present the final scores and the obtained optimal sequences as well as ground-truth action intervals.
  The horizontal axis in each plot denotes the timesteps of the video, while the vertical axis in the first plot indicates the score values ranging from 0 to 1.
  For each example, different colors correspond to different action categories, while the gray color indicates the background class.
  }
  \label{fig:sequence_visualization}
\end{figure*}

In \Fref{fig:sequence_visualization}, we visualize the obtained optimal sequences for the examples from the three benchmarks.
In the first example from THUMOS'14 (a), the optimal sequence covers the ground-truth action instances well so that the model could learn action completeness from it.
Moreover, although the examples from GTEA (b) and BEOID (c) contain a variety of action classes in a single video, our method successfully finds the optimal sequence that shows large overlaps with the ground-truth ones.
Overall, it is shown from all the examples that the optimal sequences are quite accurate even though they are selected based on point-level labels without full supervision.
They in turn provide completeness guidance to our model, which proves to improve localization performances at high IoU thresholds in \Sref{subsec:analysis} of the main paper.

\subsection{More Qualitative Comparison}
\begin{figure*}[t]
  \centering
  \includegraphics[clip=true, width=1.0\textwidth]{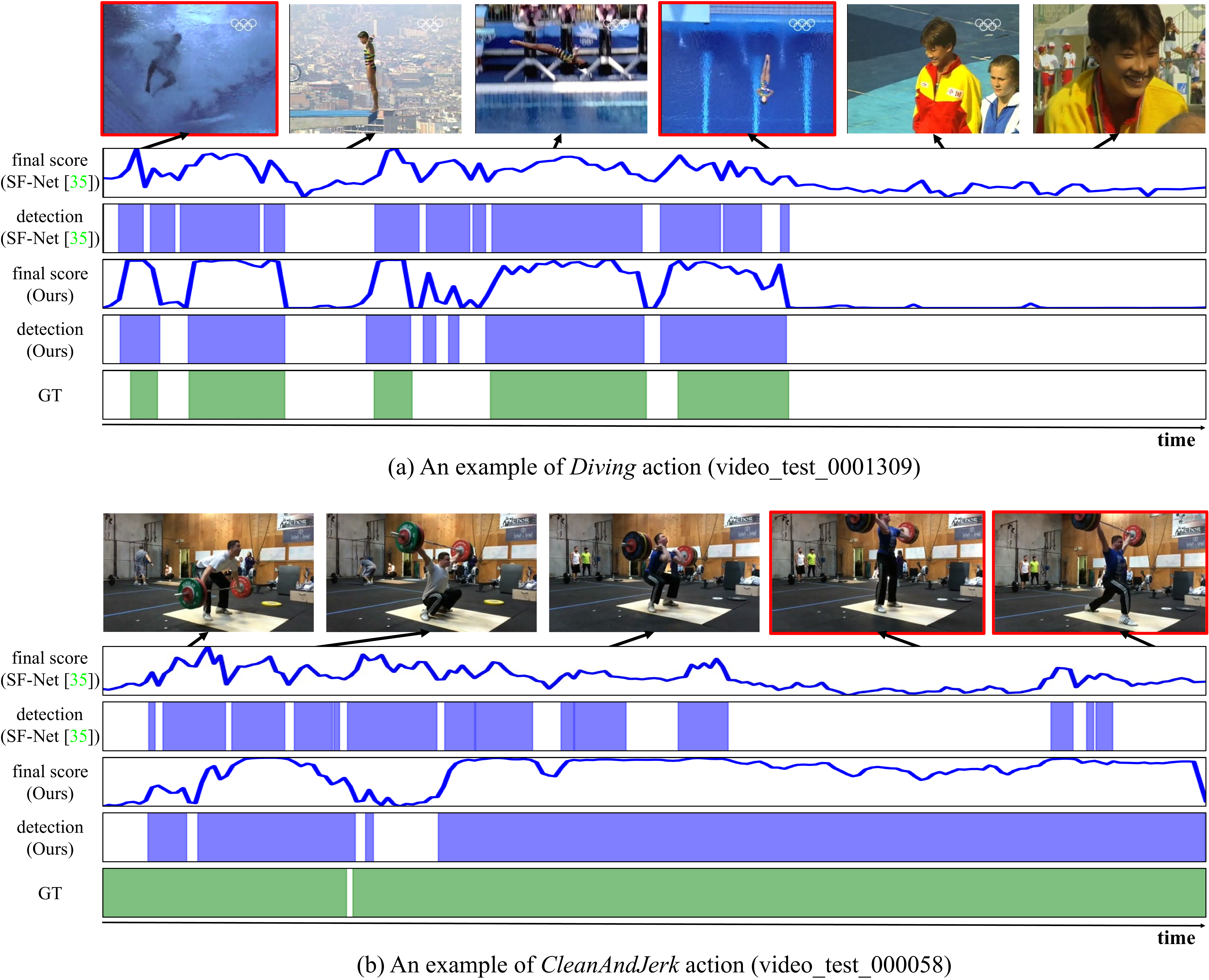}
  \caption{Qualitative comparison with SF-Net~\cite{ma2020sfnet} on THUMOS'14.
  We provide two examples with different action classes: (a) \textit{Diving} and (b) \textit{CleanAndJerk}.
  For each video, we present the final scores and detection results from SF-Net and our model as well as ground-truth action intervals.
  The horizontal axes denote the timesteps of the video, while the vertical axes are the score values ranging from 0 to 1.
  The detection threshold is set to 0.2 for our method and set to the mean score for SF-Net following the original paper.
  The red boxes indicate the frames that are misclassified by SF-Net but detected by our method.
  All of our detection results show high IoUs ($>$ 0.5) with the corresponding ground-truths regardless of their lengths.
  }
  \label{fig:qualitative_comparison_thumos}
\end{figure*}

\begin{figure*}[t]
  \centering
  \includegraphics[clip=true, width=1.0\textwidth]{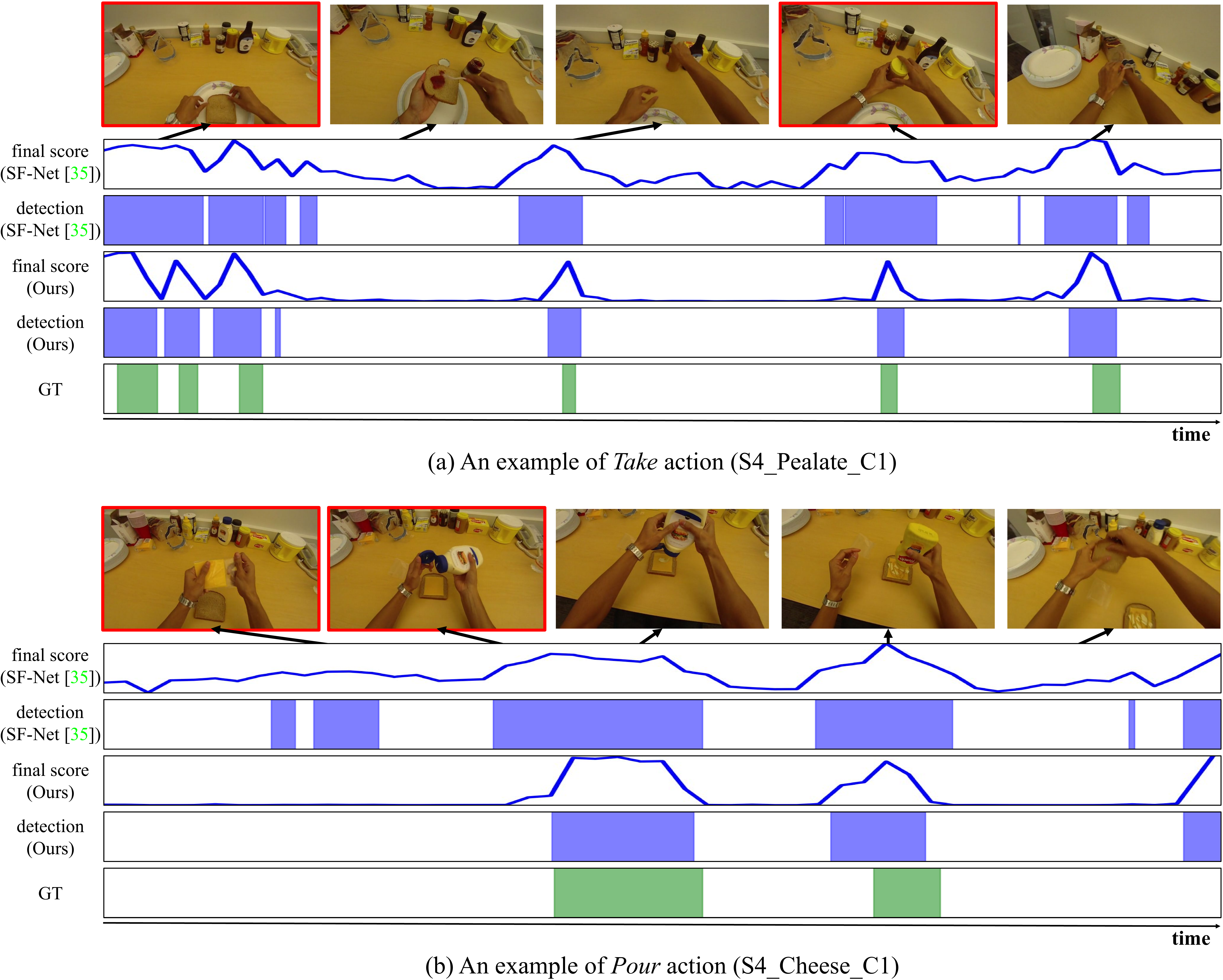}
  \caption{Qualitative comparison with SF-Net~\cite{ma2020sfnet} on GTEA.
  We provide two examples with different action classes: (a) \textit{Take} and (b) \textit{Pour}.
  For each video, we present the final scores and detection results from SF-Net and our model as well as ground-truth action intervals.
  The horizontal axis in each plot denotes the timesteps of the video, while the vertical axes are the score values ranging from 0 to 1.
  The detection threshold is set to 0.2 for our method and set to the mean score for SF-Net following the original paper.
  The red boxes indicate false alarms of SF-Net, but they, however, are rejected by our method.
  Compared to SF-Net, our method localizes action instances more precisely with fewer false positives.
  }
  \label{fig:qualitative_comparison_gtea}
\end{figure*}

\begin{figure*}[t]
  \centering
  \includegraphics[clip=true, width=1.0\textwidth]{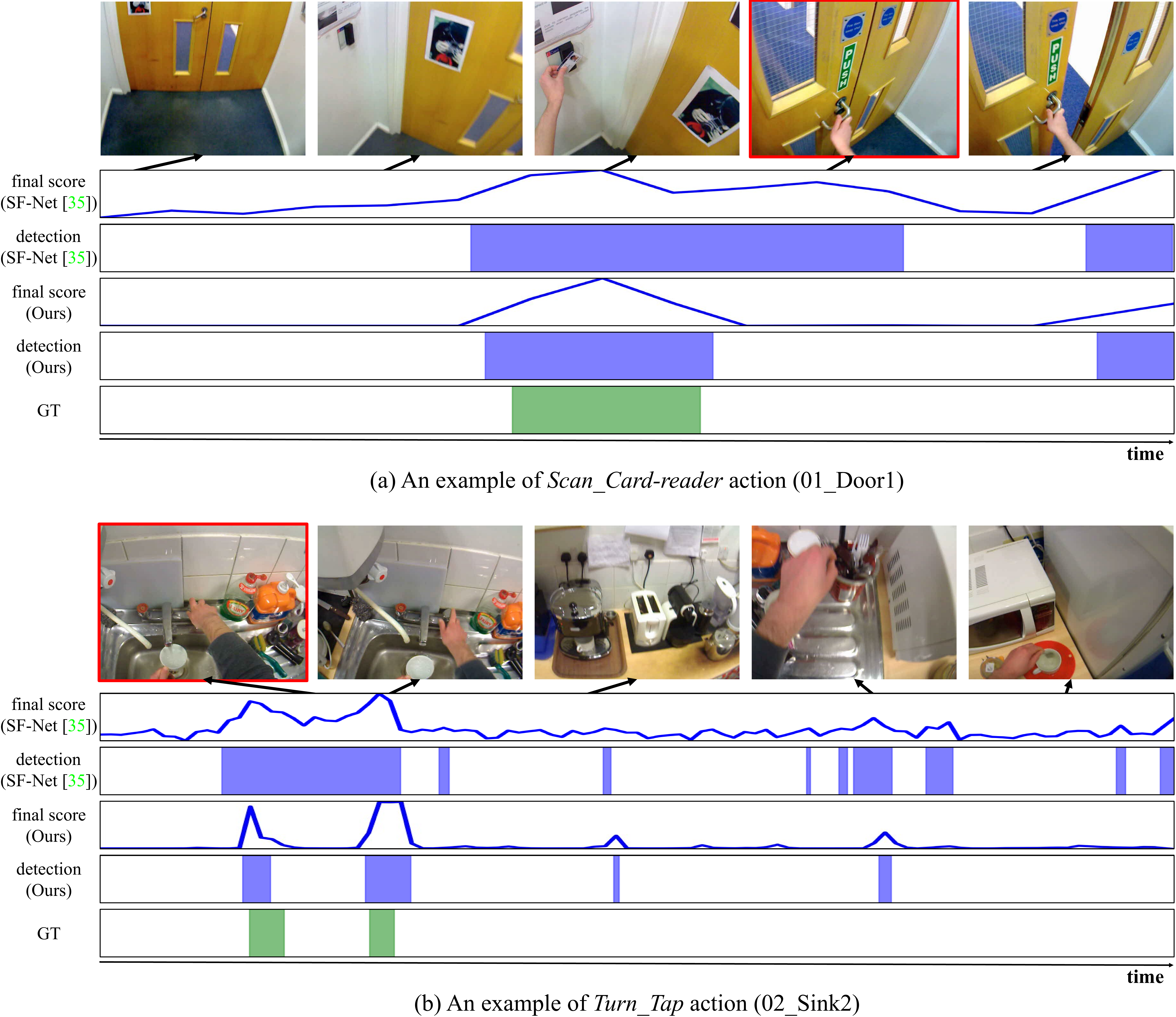}
  \caption{Qualitative comparison with SF-Net~\cite{ma2020sfnet} on BEOID.
  We provide two examples with different action classes: (a) \textit{Scan}$\underbar{~~}$\textit{Card-reader}
  and (b) \textit{Turn}$\underbar{~~}$\textit{Tap}.
  For each video, we present the final scores and detection results from SF-Net and our model as well as ground-truth action intervals.
  The horizontal axis in each plot denotes the timesteps of the video, while the vertical axes are the score values ranging from 0 to 1.
  The detection threshold is set to 0.2 for our method and set to the mean score for SF-Net following the original paper.
  The red boxes indicate false alarms of SF-Net deteriorating the performances at high IoU thresholds.
  While SF-Net overestimates the action instances, our method detects the complete action instances by discriminating action instances from background ones well.}
  \label{fig:qualitative_comparison_beoid}
\end{figure*}

We qualitatively compare our method with SF-Net~\cite{ma2020sfnet} on the three benchmarks.
The comparison on THUMOS'14~\cite{THUMOS14} is demonstrated in \Fref{fig:qualitative_comparison_thumos}.
As shown, SF-Net produces fragmentary predictions by splitting action instances, whereas our method outputs complete ones with high IoUs even for the extremely long action instance (b).
The comparison result on GTEA~\cite{lei2018gtea} is presented in \Fref{fig:qualitative_comparison_gtea}.
It would be noted that action localization on GTEA is challenging as the frames with different action categories are visually similar, leading to false positives.
We see that SF-Net has difficulty in distinguishing action instances from background ones, resulting in inaccurate localization.
On the other hand, our method successfully finds the action instances by learning completeness, showing fewer false positives.
Lastly, the comparison on BEOID~\cite{damen2014BEOID} is shown in \Fref{fig:qualitative_comparison_beoid}.
It can be clearly noticed that SF-Net fails to predict the ending times of action instances, leading to the overestimation problem.
On the contrary, with the help of the completeness guidance, our method better separates actions from their surroundings and locates the action instances more precisely.

\clearpage


\clearpage

{\small
\bibliographystyle{ieee_fullname}
\bibliography{egbib}
}

\end{document}